\documentclass{article}

\usepackage[preprint]{neurips_2025}

\usepackage{comment}

\usepackage[utf8]{inputenc} 
\usepackage[T1]{fontenc}    
\usepackage{hyperref}       
\usepackage{url}            
\usepackage{booktabs}       
\usepackage{amsfonts}       
\usepackage{nicefrac}       
\usepackage{microtype}      
\usepackage{xcolor}         
\usepackage{amsmath,amssymb,amsthm, graphicx} 
\usepackage{algorithmic}
\usepackage{multicol, multirow}
\usepackage{bbding}
\usepackage{listings}
\newcommand{\name}{\textsc{CarDiff}}

\title{Constraint-Guided Prediction Refinement via Deterministic Diffusion Trajectories}

\author{%
  ~Pantelis Dogoulis \\
  University of Luxembourg\\
  Luxembourg\\
  \texttt{panteleimon.dogoulis@uni.lu} \\
  \And
  Fabien Bernier \\
  University of Luxembourg\\
  Luxembourg \\
  \texttt{fabien.bernier@uni.lu} \\
  \AND
  Félix Fourreau\\
  University of Luxembourg, ENPC\\
  Luxembourg, Paris \\
  \texttt{felix.fourreau@enpc.fr} \\
  \And
  Karim Tit \\
  University of Luxembourg \\
  Luxembourg \\
  \texttt{karim.tit@uni.lu} \\
  \And
  Maxime Cordy \\
  University of Luxembourg \\
  Luxembourg \\
  \texttt{maxime.cordy@uni.lu} \\
}

\begin{document}

\maketitle

\begin{abstract}
Many real-world machine learning tasks require outputs that satisfy hard constraints, such as physical conservation laws, structured dependencies in graphs, or column-level relationships in tabular data. Existing approaches rely either on domain-specific architectures and losses or on strong assumptions on the constraint space, restricting their applicability to linear or convex constraints. We propose a general-purpose framework for constraint-aware refinement that leverages denoising diffusion implicit models (DDIMs). Starting from a coarse prediction, our method iteratively refines it through a deterministic diffusion trajectory guided by a learned prior and augmented by constraint gradient corrections. The approach accommodates a wide class of non-convex and nonlinear equality constraints and can be applied post hoc to any base model. We demonstrate the method in two representative domains: constrained adversarial attack generation on tabular data with  column-level dependencies and in AC power flow prediction under Kirchhoff's laws. Across both settings, our diffusion-guided refinement improves both constraint satisfaction and performance while remaining lightweight and model-agnostic.
\end{abstract}


\section{Introduction}
\label{sec:intro}
\label{intro}
Many machine learning tasks require generating predictions that are not only accurate, but also satisfy explicit structural or physical constraints. These constraints may stem from conservation laws, graph relationships, logical rules, or statistical dependencies, and define a solution space that can be casted as a lower-dimensional manifold within the ambient output space. In practice, failure to satisfy these constraints can lead to nonphysical behavior, invalid outputs, or poor generalization.

 In the context of constraint-aware learning, most existing methods fall into two distinct paradigms. The first focuses on embedding constraints into the learning process through reparameterization or projection, often targeting specific constraint types such as linear equalities or convex feasibility sets \citep{simonetto2024caa,christopher2024constrained}. These approaches have shown promise in domains like tabular data generation, but they rarely generalize to nonlinear, structured, or hybrid constraints. The second line of work, rooted in physics-informed machine learning (PIML), focuses on learning time-dependent systems governed by differential equations \citep{raissi2019physics,karniadakis2021physics}. Typically, these approaches formulate the learning task as an optimization over a dynamical system, explicitly incorporating temporal supervision and enforcing the governing boundary conditions throughout training.

In contrast, many important prediction tasks involve static systems with nonlinear constraints, where valid outputs must lie on a solution manifold defined implicitly by a set of equations. Examples include power system equilibria, mechanical force balances, and consistency rules in structured data \citep{leyli2022lips}. These settings demand methods that can refine approximate predictions into constraint-satisfying outputs, without assuming specific constraint forms or relying on time-based supervision.

In this work, we introduce a general framework for constraint-aware and model-agnostic refinement based on deterministic denoising diffusion implicit models (DDIMs) \citep{song2020denoising} that can be used to satisfy nonlinear and non-convex constraints. Starting from an approximate prediction, our method applies a learned, diffusion-inspired refinement process that progressively moves the output closer to a feasible solution. Unlike methods that encode constraints directly into the architecture or training loss, our approach operates as a post hoc projection mechanism that can be applied to any base model and any differentiable constraint formulation.

We instantiate this framework on two qualitatively different tasks. First, we consider tabular data generation under column-level constraints, such as marginal dependencies and linear totals in the scope of adversarial machine learning \citep{kurakin2016adversarial}. 
Second, we address the AC Power Flow (ACPF) problem, a canonical constrained system involving nonlinear and non-convex algebraic equations \citep{tinney1967power}.
Across both domains, \name~ significantly improves constraint satisfaction while keeping the performance close to or above the compared approaches.
\paragraph{Contributions.} In summary, we:
\begin{enumerate}
    \item Introduce \name~ (Constraint-Aware Refinement with Diffusion): a general method for constraint-aware prediction via deterministic diffusion-based refinement.
    \item Provide mathematical intuition and illustration of limitations regarding the convergence of our method on a simple didactic example.
    \item Validate \name~on two distinct real-world applications: (\textit{a}) adversarial attacks generation on tabular data with column constraint and (\textit{b}) a system described from physical equations.
\end{enumerate}

\section{Related work}
\label{sec:related}

\paragraph{Constraint Satisfaction in Generation:} Constrained generation with diffusion models has evolved rapidly. Early samplers imposed simple convex bounds through reflected or barrier processes that keep the trajectory within a feasible set \citep{fishman2023metropolis}. Subsequent research reframed every denoising step as a constrained optimization, enabling movement on far more intricate sets: projected diffusion injects an explicit projection operator \citep{christopher2024constrained}, while trust-region sampling balances a constraint-violation gradient against a noise-level “trust” metric \citep{huang2024constrained}. Orthogonal to these optimization-view methods, score guidance alters the model score with any differentiable reward, starting from classifier guidance \citep{dhariwal2021diffusion}, extending to data-consistency losses for inverse imaging \citep{chung2022improving} and to manifold preserving objectives \citep{he2024manifold}. Theoretical results shows that score networks already concentrate mass near low-dimensional manifolds, so modest corrective gradients can steer samples toward equality sets \citep{pidstrigach2022score}. When analytic projections are unavailable, practitioners employ latent-space repair: a few dozen proximal or gradient steps in the latent code progressively reduce violations while staying close to the prior \citep{huang2024constrained,zampini2025training}. At the other extreme, constraint-aware training modifies the generative architecture itself, manual diffusion bridges inject constraints into the noise schedule \citep{fishman2023diffusion}, whereas constraint layers (mainly on GAN architectures) for tabular synthesis deterministically generate outputs to user-defined rules \citep{stoian2024realistic}. These designs yield exact feasibility, but require customized retraining and often assume linear or convex constraint spaces.

\paragraph{Constraint Satisfaction in Prediction:}
In parallel, there are many works that focus on constraint satisfaction in predictive models. Physics-informed neural networks add regularization terms for the governing equations \citep{raissi2019physics}, yet they provide asymptotical satisfaction of the constraints \citep{karniadakis2021physics}. However, ongoing research has shown that these models are hard to optimize and might under-perform on more complex physical phenomena \citep{krishnapriyan2021characterizing}. Exact enforcement appears either inside differentiable optimization layers serving as quadratic program solvers \citep{amos2017optnet}, cone projections \citep{agrawal2019differentiable}, or in unrolled correction modules such as DC3, which projects intermediate predictions during training \citep{donti2021dc3}. However, the latter method requires heavy memory and big computation time and it cannot be applied post-hoc to any coarse approximation of the problem. Test-time optimization frameworks perform a similar repair after the predictor fires, for example in decision-focused "predict-and-optimize" pipelines \citep{wilder2019melding}. Finally, the authors in \cite{min2024hard} propose HardNet, a neural-network architecture that appends a differentiable projection layer so every forward pass exactly satisfies affine or more general convex, input-dependent constraints; they further prove the augmented network retains universal approximation power, showing hard feasibility need not trade off expressiveness. While effective, these approaches depend on problem-specific solvers and often assume convexity or inequality slack variables, leaving coupled nonlinear equalities largely untreated.

Table \ref{tab:constraint_methods} provides a comparative overview of leading constraint-aware techniques, detailing each method’s task modality, post-hoc applicability, convexity assumptions, supported constraint forms, and key operational features. Although projected diffusion samplers \citep{christopher2024constrained} and trust-region diffusion \citep{huang2024constrained} could in principle be adapted to refine an existing guess, their pipelines are optimized for fully generative sampling. A direct empirical head-to-head comparison would require nontrivial modifications and risks conflating generation quality with projection efficiency.

Our proposed method unifies ideas from projection, gradient guidance and decision-focused repair within a single deterministic DDIM trajectory. Each denoising step injects a constraint gradient, mimicking an incremental projection yet remaining lightweight, due to the nature of  DDIM, and solver-free. As a result, the algorithm \textit{(i)} enforces broad, smooth, nonlinear and possibly non-convex equalities post-hoc, \textit{(ii)} preserves sample realism through the learned diffusion prior, and \textit{(iii)} wraps around any pretrained generator or predictor without re-training or custom optimization code. In doing so, it fills a gap between inference-time guidance, which may drift off-manifold, and architecture-level designs that sacrifice plug-and-play flexibility. 

\begin{table}[t]
    \centering
    \footnotesize
    \setlength{\tabcolsep}{2pt}
    \renewcommand{\arraystretch}{0.95}
    \caption{Landscape of constraint-aware methods. “Post-hoc” denotes training-free refinement applicable to any frozen model. $^{\dagger}$\cite{he2024manifold} show empirical success on moderate manifolds; robustness to complex physics constraints is unverified.}
    
    \begin{tabular}{p{2.5cm} p{1.0cm} p{1.2cm} p{1.2cm} p{1.5cm} p{5.5cm}}
        \toprule
        \textbf{Method} & \textbf{Task style} & \textbf{Post-hoc} & \textbf{Convex} & \textbf{Linear} & \textbf{Key features}\\
        \midrule
        \cite{christopher2024constrained} & Gen. & \Checkmark &  \XSolidBrush &  non-linear eq/ineq & Requires a differentiable projector or iterative inner solve at each step.\\
        \cite{huang2024constrained} & Gen. & \Checkmark &  \XSolidBrush &  non-linear eq/ineq & Training-free; moves along constraint gradient until a trust-region bound (noise-level) is hit.\\
        \cite{he2024manifold} & Gen. & \Checkmark & Limited$^{\dagger}$ &  non-linear eq & Score guidance with small step size; scalability on highly curved manifolds untested.\\
        \cite{zampini2025training} & Gen. & \Checkmark &  \XSolidBrush &  non-linear eq/ineq & Outer-loop proximal or gradient steps on latent code; training-free but slower.\\
        \cite{fishman2023diffusion} & Gen. & \XSolidBrush &  \XSolidBrush &  non-linear & Custom architecture; constraints embedded in the noise schedule.\\
        \cite{stoian2024realistic} & Gen. & \XSolidBrush &  \Checkmark &  linear / rule & Tabular GANs; deterministic snapping to rule set.\\
        \midrule
        \cite{donti2021dc3} & Pred. & Both &  \XSolidBrush &  non-linear eq/ineq & Unrolled differentiable projector; heavy memory.\\
        \cite{min2024hard} & Pred. & Arch. &  \Checkmark &  affine only & Projection layer with universal approximation proof.\\
        \cite{amos2017optnet} / \cite{agrawal2019differentiable} & Pred. & During training &  \Checkmark &  linear/quad & Differentiable convex solver inside network.\\
        \midrule
        \name~(\textbf{ours}) & Ref. & \Checkmark &  \Checkmark & non-linear / linear & Constraint-aware and model-agnostic method to refine original predictions.\\
        \bottomrule
    \end{tabular}
    \label{tab:constraint_methods}
\end{table}

\section{Method}
\label{sec:method}
\subsection{Preliminaries}

\paragraph{Diffusion models.}
In the context of diffusion generative modeling, one defines a forward Markov process that gradually corrupts an arbitrary clean sample $x\sim p(X)$ by successively adding isotropic Gaussian noise \citep{NEURIPS2020_4c5bcfec}. At time step \(t\), the noised sample is given by
$
  x_t 
  = 
  \sqrt{\bar\alpha_t}\,x 
  + 
  \sqrt{1-\bar\alpha_t}\varepsilon,$
where \(\{\bar\alpha_t\}_{t=1}^T\) is a pre-specified noise schedule. A neural network $\varepsilon_\theta(x_t,t)$ is then trained to approximate the true noise $\varepsilon$ added at each step.

At inference time, samples are generated by approximately reversing this diffusion chain. For Denoising Diffusion Probabilistic Models (DDPMs), this involves a stochastic process where random noise is added at each reverse step. In contrast, Denoising Diffusion Implicit Models (DDIMs) \citep{song2020denoising} use a deterministic sampling procedure that improves sampling efficiency and controllability. Given a noisy input $x_t$, one computes:
\begin{equation}
  x'_{t-1}
  = 
  \sqrt{\bar\alpha_{t-1}}\,
  \frac{x_t 
        - 
        \sqrt{1-\bar\alpha_t}\,\varepsilon_\theta(x_t,t)}
       {\sqrt{\bar\alpha_t}}
  \;+\;
  \sigma_t\,\varepsilon_t,
  \quad
  \varepsilon_t\sim\mathcal{N}(0,\text{I})
\label{eq:DDIM}
\end{equation}

where
\begin{equation}
  \sigma_t
  =
  \eta
  \sqrt{\frac{1-\bar\alpha_{t-1}}{1-\bar\alpha_t}}
  \sqrt{1-\frac{\bar\alpha_t}{\bar\alpha_{t-1}}}
\end{equation}
Setting \(\eta=1\) recovers the original stochastic DDPM sampler, in which each reverse step introduces additional Gaussian noise, whereas \(\eta=0\) yields a deterministic DDIM trajectory that traverses a non-Markovian but noise-free path through the data manifold.

\subsection{Constraint Manifold Setup and Initial Approximation}

We present a method to project approximate solutions onto constraint manifolds implicitly defined as the zero locus of a non-negative function \( \Phi:\mathbb{R}^d \to \mathbb{R}_+ \), which is differentiable and possibly non-convex. Our method combines three key components: (1) a supervised coarse estimator, (2) a denoising diffusion implicit model (DDIM) trained to guide iterates toward feasible regions, and (3) a proximal correction step using constraint gradients with geometrically meaningful step sizes.

Let the constraint manifold be defined as
\begin{equation}
    \mathcal{M} = \{ x \in \mathbb{R}^d | \Phi(x) = 0 \}
\end{equation}

and suppose we are given a noisy or unconstrained estimate \( x_{\text{init}} \in \mathbb{R}^d \). Our goal is to find a point \( x_0 \in \mathcal{M} \) that is as close as possible to \( x_{\text{init}} \), i.e.,
\begin{equation}
    x_0 \in \arg\min_{x \in \mathcal{M}} \|x - x_{\text{init}}\|^2
\end{equation}
whenever such a minimizer \textit{exists}. While this is not the case in general for non-convex constraint functions, in practice and for many cases of interest, e.g. when $\mathcal{M}$ represents stable equilibria of a physical system, a minimizer exists and is often unique.

As a first step, we train a coarse estimator $f_\theta$ to map inputs $x$ to approximate feasible solutions. This function is optimized via the usual supervised regression loss $\min_\theta \mathbb{E}_{(x, y)} \|f_\theta(x) - y\|^2$ and provides an initial approximation $x_T := f_\theta(x_{\text{init}})$, which serves as the starting point for our refinement procedure.

\subsection{\name~Prediction Refinement}

To guide predictions toward the manifold $ \mathcal{M}$, we train a diffusion model to refine predictions through a sequence of intermediate states. Define a noise schedule $ \lambda_1, \dots, \lambda_T \in (0, 1) $ and associated scalars
$\alpha_t = 1 - \lambda_t$, where $\bar{\alpha}_t = \prod_{s=1}^t \alpha_s$. 
At inference, starting from a noisy estimation $x_t$, the model deterministically maps each state backward ($x'_{t-1}$) using the DDIM sampling scheme described in Eq \ref{eq:DDIM} with $\eta = 0$.
This process encodes a smooth trajectory towards the manifold, implicitly guided by learned geometric priors from training data \citep{he2024manifold, pidstrigach2022score}. 
We augment the DDIM transition with a constraint-driven displacement inspired by \citep{dhariwal2021diffusion}. Formally, we express our update step as:
\begin{equation}
x_{t-1} = x_{t-1}' + \gamma_t\bold{\delta}_t
\end{equation}
where $\gamma_t$ is the gradient update step size and $\bold{\delta}_t=\frac{-\nabla_{x}\Phi(x_{t-1}')}{||\nabla_{x}\Phi(x_{t-1}')||_2}$ is the descent direction of $\Phi$ at $x_{t-1}'$.

\paragraph{Choosing $\gamma_t$.} At each iteration, the score network proposes a denoising direction that points towards a global estimate $\widehat{x}_0$, derived from the DDIM trajectory. Meanwhile, the constraint gradient descent direction $-\delta_t$
encodes the steepest descent path for the potential $\Phi$. We combine these two sources of information by tilting the reverse diffusion path along $\delta_t$. We propose to determine the step size $\gamma_t$ by solving a proximal objective that balances two terms: fidelity to the approximate DDIM target $\widehat{x}_0$ and approximate constraint satisfaction. Specifically, we minimize the following proximal objective:
\begin{equation}
\label{eq:gamma-objective}
\mathcal{L}(\gamma;\lambda, x_{t-1}') = \|x_{t-1}' + \gamma \delta_t - \widehat{x}_0\|^2 + \lambda \left( \Phi(x_{t-1}') + \gamma \nabla \Phi(x_{t-1}')^\top \delta_t \right)^2,
\end{equation}
where the second term corresponds to a first-order Taylor approximation of $\Phi(x)$ around $x_{t-1}'$, squared to yield a \textit{scale-invariant} penalty. Letting $r_t := \widehat{x}_0 - x_{t-1}'$, and using that $\nabla \Phi(x_{t-1}')^\top \delta_t = -\|\nabla \Phi(x_{t-1}')\|$, we obtain the closed-form minimizer:
\[
\gamma_t = \frac{ \langle r_t, \delta_t \rangle + \lambda\, \Phi(x_{t-1}')\cdot \|\nabla \Phi(x_{t-1}')\| }{1 + \lambda \|\nabla \Phi(x_{t-1}')\|^2}.
\]

To interpret this expression geometrically, we define the cosine of the angle between the DDIM correction and the constraint direction $\cos\theta_t$ and the distance $d_t$ to the current target approximation $\widehat{x}_0$: $
\cos\theta_t := \frac{\langle  r_t, \delta_t \rangle}{\|r_t\|}, d_t := \|r_t\|.
$

This yields a more interpretable form:
\[
\boxed{
\gamma_t = \frac{ d_t \cos\theta_t + \lambda\, \Phi(x_{t-1}')\cdot \|\nabla \Phi(x_{t-1}')\| }{1 + \lambda \|\nabla \Phi(x_{t-1}')\|^2}
}
\]

This expression balances the influence of global guidance (via $\widehat{x}_0$) with the local constraint violation (via $\Phi(x_{t-1}')$ and $\nabla \Phi(x_{t-1}')$). When the DDIM prediction already aligns with the constraint direction (i.e., $\cos\theta_t > 0$), the gradient correction is increased and, conversely, when they disagree, the gradient correction is softened. Additionally, when the constraint violation dominates, the proximal term  adjusts the step size accordingly, but with a scale-controlled behavior due to the  denominator. 
Mostly, in practice, we notice that $\lambda \approx 0$, which is equivalent to taking $\gamma_t \propto d_t \cos\theta_t $, provides good results.

\paragraph{Connection with PINN loss.} Although our method does not fit the standard definition of a Physics-Informed Neural Network (PINN) \citep{raissi2019physics}, each guided diffusion step can be interpreted as solving in closed form, a single-sample instance of the canonical PINN objective:
\begin{equation}
    \min_{x} \underbrace{||x - \widehat{x}_0||^2}_\text{data prior} + \underbrace{\lambda \Phi(x)^2}_{\text{physics term}}
\end{equation}
with all network parameters held \textit{fixed}. In this sense, the diffusion prior supplies the empirical component while the constraint potential plays the role of the physics residual, yielding a sample-wise, one-step analogue of the optimization that a PINN performs over its  parameter space.

\section{Illustration of the Proposed Method}
\label{sec:illustration}
To illustrate some properties of the proposed method, we consider the classical Müller–Brown potential \citep{brown1971reversal}, a widely-used function representing a two-dimensional energy landscape in molecular modeling. It describes the energy of a molecular system as a function of its parameters projected in 2 dimensions and can be expressed as:

\begin{equation}
    \Phi(x, y) = \sum_{i=1}^4 A_i \exp\!\left[ a_i (x - x_i)^2 + b_i (x - x_i)(y - y_i) + c_i (y - y_i)^2 \right]
\end{equation}

where \( (x, y) \) represent simplified reaction parameters of the system, \( A_i \) control the depth of each potential well, and \( (x_i, y_i) \) specify their centers. The parameters \( a_i \), \( b_i \), and \( c_i \) determine the curvature and orientation of each component. For all numerical experiments, we utilize the canonical parameter set (see Section C of the Appendix), which yields a surface comprising three distinct minima and a single saddle point. 


\begin{figure}[h]
    \centering
    \includegraphics[width=0.98\linewidth]{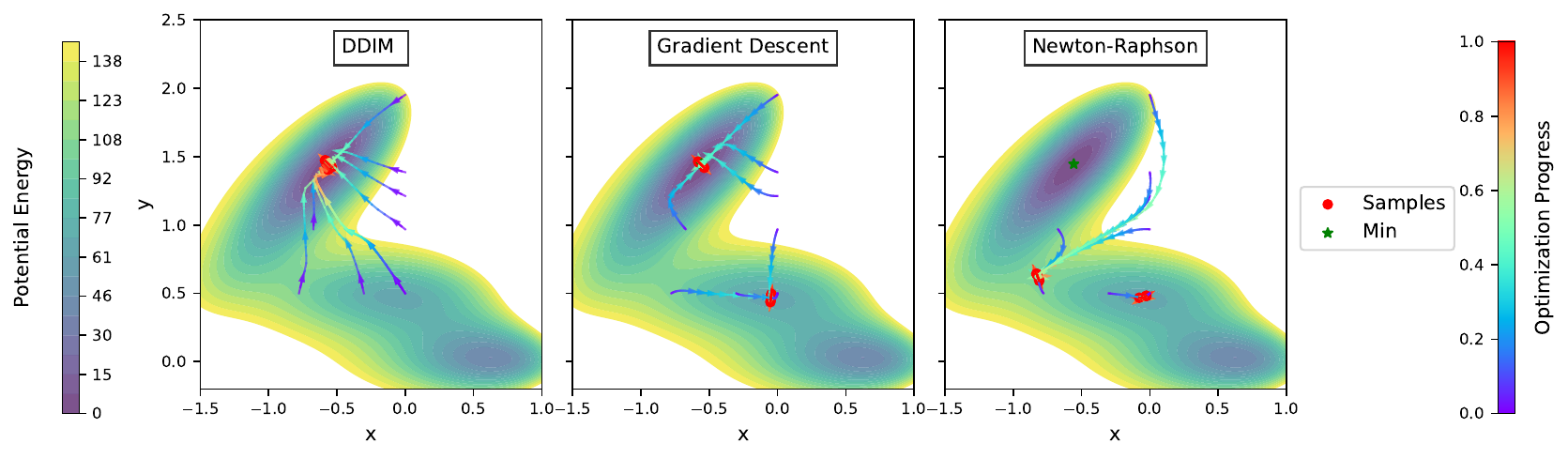}
    \caption{Optimization trajectories on the Müller–Brown potential with the same initial points.}
    \label{fig:toy-example-results}
\end{figure}

In Figure~\ref{fig:toy-example-results}, we highlight the limitations of conventional optimization techniques. Gradient descent frequently converges to local minima, while the Newton–Raphson method often fails by converging to the saddle point, largely due to the Hessian, which does not distinguish between minima and saddles. In contrast, the proposed diffusion-based approach consistently reaches the global minimum, with the progressive denoising steps enabling the model to bypass local minima early in the optimization process. 

A principal limitation of our approach arises when the initial sample $x_T$ lies too far from the true data manifold. In such cases, the predetermined diffusion depth $T$ may be insufficient to traverse the high-energy barriers separating the sample from the locally convex region enclosing the global minimum. Consequently, subsequent guided updates tend to drive the iterates into the nearest shallow well, yielding convergence to a suboptimal local minimum instead of the desired global optimum (see Figure \ref{fig:fail}). The detailed implementation can be found in Section C of the Appendix.
\begin{figure}[h]
    \centering
    \includegraphics[width=0.65\linewidth]{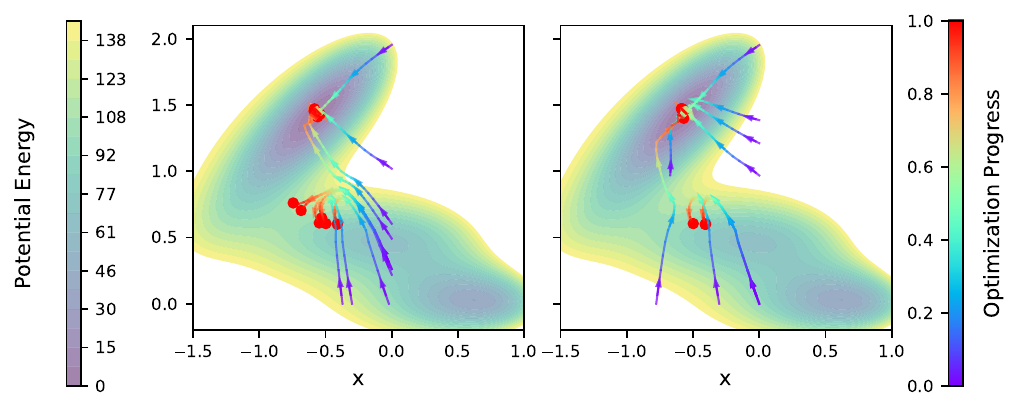}
    \caption{Two different cases of diffusion based trajectories failing to reach the global minimum.}
    \label{fig:fail}
\end{figure}

\section{Experiments}
\label{sec:experiments}
\subsection{Constrained Adversarial Attacks}
\subsubsection{Problem Formulation}
Adversarial examples are inputs to a machine learning model that have been intentionally perturbed to cause the model to make incorrect predictions. The perturbation is typically small, but large enough to mislead the model. Formally, given an input  $x$ and a model $f(x)$, an adversarial example $x'$ is defined as: $x' = x + \delta$ , where $\delta$ is a perturbation added to $x$, and $\| \delta \|$ is typically constrained to be small, often measured by a norm like $\|\delta\|_{\infty} \leq \epsilon$, where $\epsilon$ is a small constant.

For constraint-based adversarial attacks, the goal is to generate adversarial examples while ensuring that they satisfy some predefined constraints. This can be formulated as a constrained optimization problem \citep{ijcai2022p183,simonetto2024caa}. The general approach to such attacks involves computing the gradient of the loss function with respect to the input $x$, i.e., $\nabla_x L(x)$. After computing the gradient, the attack aims to adjust $x$ in the direction of the gradient, while ensuring that the resulting perturbed input remains feasible under the given constraints, which are encoded in $L$. This adjustment is typically done through iterative methods, where at each step, the input is modified by adding a small perturbation in the direction of the gradient, and the feasibility is checked or enforced using some procedure. The update rule can be generally expressed as $x' = x + \eta \cdot \text{sign}(\nabla_x L(x))$, where $\eta$ is the step size. 

\paragraph{Cyclic Refinement on PGD}
We integrate our refinement process into the standard PGD attack by alternately applying PGD updates and DDIM projections. 
Given a clean input $x^{(0)}$ and true label $y$, we alternate for $C$ cycles between: (\textit{i}) $k$ PGD steps 
and (\textit{ii}) the proposed diffusion-based projection via $\tau$ DDIM reverse steps where we apply a gradient descent step on the constraint potential $\Phi$.
We can summarize the sequence of operations in the following diagram:
$$
\underbrace{\mathrm{PGD}}_{\text{adversarial perturbation}}
\longrightarrow
\underbrace{\mathrm{DDIM\ refinement}}_{\substack{\text{projection}\\\text{onto constraints}}}
\longrightarrow
\underbrace{\mathrm{PGD}}_{\text{adversarial perturbation}}
\longrightarrow
\cdots
\longrightarrow\;
\underbrace{\mathrm{DDIM\ refinement}}_{\substack{\text{projection}\\\text{onto constraints}}}
$$

\subsubsection{Setup and results}

\paragraph{Experimental setup.}
We evaluate four adversarial attack methods: standard PGD \citep{madry2018towards}, CPGD \citep{ijcai2022p183}, CAPGD \citep{simonetto2024caa}, and our proposed cyclic refinement in a tabular classification task. In this section, all the experiments are repeated 5 times on a computer with GPU \textsc{NVIDIA GeForce RTX 4080 SUPER}. In addition, experiments are conducted on the \texttt{lcld\_v2\_iid} dataset from TabularBench \citep{simonetto2024tabularbench}, which is a credit-scoring dataset and includes mixed numerical and categorical features with predefined lower/upper bounds and relational constraints. Based on this, we can express the constraint potential $\Phi(x) = \sum_{r=1}^6 \bigl(c_r(x)\bigr)^2$ as a sum of squared violations of the relational constraints. The exact formulas corresponding to each constraint function $c_1,\cdots,c_6$ can be found in Section A of the Appendix.

 \paragraph{Evaluation Metrics.} We report the Robust Accuracy (a widely used metric in the domain of adversarial learning) and the constraints violation of the potential function $\mathbf{\Phi}$.

 \paragraph{Constraint satisfaction effectiveness.} As shown in Table \ref{tab:adv-res}, \name~achieves the lowest constraint violation, with the next-best approach exhibiting roughly an order of magnitude larger violation, while still preserving robust accuracy comparable to that of standard PGD. This demonstrates that our cyclic refinement produces adversarial examples that are equally effective at fooling the network but adhere far more closely to the data constraints, compared with the the current state‐of‐the‐art constrained attack methods found in the literature.

\begin{table}[ht]
    \centering
    \caption{Comparison of adversarial attack methods on Robust Accuracy and Constraint Violation ($\pm$ standard error). Best performance is in \textbf{bold} while the second best is \underline{underlined}.}
    \label{tab:results}
    \begin{tabular}{lcc}
        \toprule
        \textbf{Attack} & $\downarrow$ \textbf{Robust Accuracy (\%)} & $\downarrow$ \textbf{Constraint Violation} ($\mathbf{\Phi}$) \\
        \midrule
        PGD   & 46.608 \scriptsize{$\pm0.009$} & 29.931 \scriptsize{$\pm 0.041$} \\
        CPGD  & 51.386 \scriptsize{$\pm0.010$} & \underline{14.282} \scriptsize{$\pm0.027$} \\
        CAPGD & \textbf{30.828} \scriptsize{$\pm0.003$} & 157.765 \scriptsize{$\pm0.120$} \\
        \name~& \underline{39.769} \scriptsize{$\pm 0.011$} & \textbf{1.841} \scriptsize{$\pm 0.003$}\\
        \bottomrule
    \end{tabular}
\label{tab:adv-res}
\end{table}

\subsection{AC Power Flow Prediction}
\subsubsection{Problem Formulation}
A power grid is an interconnected electrical network whose nodes (buses) represent generators, loads, and substations, while its edges (transmission lines) convey power between those buses. The power flow prediction problem requires determining the voltage magnitudes \( |V_i| \) and phase angles \( \theta_i \) at each bus \( i \) given known active and reactive power injections, \( P_i \) and \( Q_i \). This is mathematically formulated through the nonlinear algebraic alternative current (AC) power flow equations, widely known as Kirchhoff's power flow equations. 


\begin{equation}
\begin{cases}
P_i = V_i \sum_{j=1}^{N} V_j \left( G_{ij} \cos\theta_{ij} + B_{ij} \sin\theta_{ij} \right) 
\\
Q_i = V_i \sum_{j=1}^{N} V_j \left( G_{ij} \sin\theta_{ij} - B_{ij} \cos\theta_{ij} \right) \end{cases}
\label{eq:KCL}
\end{equation}

The power flow equations are inherently non-convex due to the presence of trigonometric functions in the voltage angle differences. In theory, a given set of active and reactive power injections can yield multiple solutions, however not all of them are physically meaningful. 

The non-convex nature of the problem further implies that standard optimization techniques might converge to local solutions rather than a unique global one. Although iterative methods such as the Newton-Raphson algorithm are commonly used and typically converge to the physical solution when provided with a good initial guess, the existence of multiple mathematical solutions is a well-documented phenomenon in power system analysis \citep{akram2015newton}. Generally, among the multiple feasible solutions, only one is robust in terms of stability (also called the stable operating point), while the others are either not observable in normal operating conditions or are indicative of system collapse scenarios.


\subsubsection{Setup and results}
\paragraph{Experimental setup.} In our evaluation of the proposed constraint-guided refinement method, we conducted extensive experiments on AC power flow prediction problems of varying complexity.
We evaluated our approach on two standard IEEE test cases: the 14-bus and 30-bus power systems. Each dataset comprises a single grid instance with nominal values assigned to every bus, containing 14 and 30 buses, respectively. To adapt these datasets for a machine learning framework, we augment the available data by generating multiple independent and identically distributed (iid) instances by sampling; (\textit{i}) with a narrow distribution near the nominal values (for training), (\textit{ii}) with a wider distribution centered on the nominal values (for testing), to evaluate robustness on different operational scenarios (details can be found in the Section B of the Appendix). For each system, we generated datasets consisting of 10,000 training samples, 5,000 validation samples, and 10,000 test samples, and the experiments are repeated 5 times.
All ground truth solutions were computed using the Newton-Raphson method, which serves as our numerical baseline for constraint satisfaction.

Our experiments compare three approaches:
\begin{itemize}
    \item \textbf{Baseline (PINN)}: A physics-informed fully connected neural network that incorporates power flow equations directly into its loss function \citep{9303008};
    \item \textbf{Base Estimator}: Same architecture, but trained to predict power flow solutions without explicit constraint enforcement;
    \item \textbf{Diffusion Refined}: Our proposed method \name~ that applies constraint-guided diffusion to refine
the base estimator's predictions. 
\end{itemize}

For the diffusion model, we used a simplified U-Net architecture with a noise schedule optimized on the validation set. Further details about the implementation  and the hyperparameters can be found in Sec. B of the Appendix.

\paragraph{Evaluation Metrics.} In addition to the Mean Squared Error, we measure the Maximum Active and Reactive Power Mismatch (resp. denoted MAPM, MRPM), which represents the maximum violation of active and reactive power balance constraints for each sample. Reporting the maximum violation clearly captures the worst-case imbalance, which is a substantially important metric for real-world power grid operators in applications like Voltage Control \citep{olival2017advanced,murray2021voltage}. 

\begin{table}[!ht]
\centering
\caption{Comparison of the power flow prediction methods on MSE and Constraint Violation ($\pm$ standard error). Best performance is in \textbf{bold} while the second best is \underline{underlined}.}
\begin{tabular}{l l r r r}
\toprule
\textbf{Case} & \textbf{Model} & \multicolumn{1}{c}{$\downarrow$ \textbf{MSE}} & \shortstack{Constraint 1 \\ $\downarrow$ \textbf{MAPM} ($\text{MW}^2$)} & \shortstack{Constraint 2 \\ $\downarrow$ \textbf{MRPM} ($\text{MVAr}^{2}$)} \\
\hline
\multirow{3}{*}{case14} 
    & Baseline (PINN) & $\textbf{0.648}$ {\scriptsize $\pm 0.0014$} & $\underline{19.2913}$ {\scriptsize $\pm 0.1015$} & $\underline{31.99}$ {\scriptsize $\pm 0.1902$} \\
    & Base Estimator & $0.666$ {\scriptsize $\pm 0.0011$} & $111.15$ {\scriptsize $\pm 0.5756$} & $327.1$ {\scriptsize $\pm 1.6961$} \\
    & \name & $\underline{0.665}$ {\scriptsize $\pm 0.0012$} & $\textbf{14.51}$ {\scriptsize $\pm 0.0811$} & $\textbf{20.02}$ {\scriptsize $\pm 0.1209$} \\
\hline
\multirow{3}{*}{case30} 
    & Baseline (PINN) & $4.660$ {\scriptsize $\pm 0.0254$} & $\underline{21.31}$ {\scriptsize $\pm 0.1528$} & $\underline{51.87}$ {\scriptsize $\pm 0.4507$} \\
    & Base Estimator & $\textbf{2.935}$ {\scriptsize $\pm 0.0160$} & $73.49$ {\scriptsize $\pm 0.5316$} & $212.2$ {\scriptsize $\pm 1.4520$} \\
    & \name & $\underline{3.204}$ {\scriptsize $\pm 0.0162$} & $\textbf{13.04}$ {\scriptsize $\pm 0.0698$} & $\textbf{29.57}$ {\scriptsize $\pm 0.1925$} \\
\bottomrule
\end{tabular}
\label{tab:acpf}
\end{table}

\begin{table}[!ht]
    \centering
    \caption{Average run times for ACPF predictions (milliseconds)}
    \begin{tabular}{l r r}
        \toprule
        \textbf{Model} & \textbf{case14} & \textbf{case30} \\
        \hline
        Baseline (PINN)      & $0.0101$ & $0.0108$ \\
        Newton-Raphson       & $14.1578$ & $14.9582$ \\
        \name & $4.5370$ & $4.6149$ \\
        \bottomrule
    \end{tabular}
    \label{tab:runtimes}
\end{table}

\paragraph{Constraint satisfaction effectiveness.} As shown in Table \ref{tab:acpf}, our diffusion refinement method significantly improves constraint satisfaction over the base estimator, reducing MAPM by 87\% for case14 and 82\% for case30. Similarly, the second constraint, MRPM, shows substantial improvements of 94\% and 86\% respectively.
These gains come with only 10\% increase in MSE for case30, which is still 32\% less than the PINN model's. For case14, \name~shows no MSE degradation, while dramatically improving constraint satisfaction.

\paragraph{Computational efficiency.} From a computational perspective, our approach presents an attractive middle ground between the baseline PINN model and the traditional Newton-Raphson method (see Table \ref{tab:runtimes}). While pure Newton-Raphson achieves perfect constraint satisfaction, it requires from 14 to 15ms per instance. In contrast, our diffusion refinement method takes only about 8.2ms per instance while achieving constraint satisfaction levels that are similar with fast approaches used from real-world power grid operators --- e.g. DCPF method found in \cite{seifi2011electric}. The code can be found in the Supplementary Material and will be released after the end of the review process.

\section{Limitations and Broader Impact}
\label{sec:lim}
\paragraph{Limitations.} A key limitation of our proposed refinement method is its heavy reliance on both the quality of the initial guess and the differentiability of the constraints function. First, the DDIM-based trajectory is more effective when the initial estimate resides to a point towards the data manifold (see Figure \ref{fig:fail} in Section \ref{sec:illustration}). Secondly, every corrective step requires the gradient $g_t$, so the method explicitly assumes that $\mathbf{\Phi}$ is a smooth function. Discontinuous, combinatorial, or rule-based constraints cannot be enforced without ad-hoc smoothing or external projection steps, introducing additional approximation errors.

\paragraph{Broader Impact.}
In power‐system applications, the proposed method offers a surrogate for Newton-based solvers or a refinement procedure for AI-based models, enabling accurate and faster predictions. Furthermore, our method can be also utilized in the \textit{Reinforcement Learning} community, by enabling agents to obtain faster grid-state estimates, thereby accelerating research on safety-constrained control and planing of power systems \citep{chen2021powernet,meng2024online}.
In adversarial settings however, constraint-aware refinement is twofold: Classifier training can be augmented by adversarial examples generated from our method, resulting into an improved robustness. However, the same mechanism can be exploited to craft constraint-preserving attacks.

\section{Conclusion}
\label{sec:conclusion}
In this work, we introduce \name, a model-agnostic and constraint-aware refinement procedure that injects explicit gradient guidance into a deterministic DDIM reverse trajectory. The resulting operator acts as a fast and accurate projection that can be anchored to any base predictor whenever the governing constraints are differentiable and available in analytical form.  A didactic study in Sec. \ref{sec:illustration} showcased the convergence behavior of our method, and in Sec. \ref{sec:method} we showcased the mathematical intuition and the geometric role of the cosine-scaled step size. Furthermore, \name~produces constraint-satisfactory results in both non-convex and nonlinear domains as shown in Sec. \ref{sec:experiments}. Finally, future work includes the convergence analysis of \name~under specific assumptions on the constraint function and empirical evaluations on other tasks of constraint-aware literature.
\bibliographystyle{plainnat} 
\bibliography{bibtex} 

\end{document}